\documentclass{article} % For LaTeX2e
\usepackage{nips15submit_e,times}
\usepackage{hyperref}
\usepackage{times}
\usepackage{helvet}
\usepackage{courier}
\usepackage{amssymb}
\usepackage{latexsym}
\usepackage{amsmath}
\usepackage{multirow}
\usepackage{url}
\usepackage{algorithm}
\usepackage{algpseudocode}
\usepackage{amsmath}
\usepackage{rotating}
\usepackage{epsf}
\usepackage{enumitem}
\title{Strategic Dialogue Management via Deep Reinforcement Learning}
\author{
Heriberto Cuay\'ahuitl \\
Interaction Lab \\ Department of Computer Science\\
Heriot-Watt University\\
Edinburgh \\
\texttt{ hc213@hw.ac.uk} \\
\And
Simon Keizer \\ Interaction Lab \\ Department of Computer Science\\
Heriot-Watt University\\
Edinburgh \\
\texttt{s.keizer@hw.ac.uk} \\
\AND
Oliver Lemon \\ Interaction Lab \\ Department of Computer Science\\
Heriot-Watt University\\
Edinburgh \\
\texttt{o.lemon@hw.ac.uk} \\
}

\nipsfinalcopy % Uncomment for camera-ready version

\begin{document}

\maketitle

\begin{abstract}
Artificially intelligent agents equipped with strategic skills that can negotiate during their interactions with other natural or artificial agents are still underdeveloped. This paper describes a successful application of Deep Reinforcement Learning (DRL) for training intelligent agents with strategic conversational skills, in a situated dialogue setting. Previous studies have modelled the behaviour of strategic agents using supervised learning and traditional reinforcement learning techniques, the latter using tabular representations or learning with linear function approximation. In this study, we apply DRL with a high-dimensional state space
%OL : is it really continuous?
 to the strategic board game of Settlers of Catan---where players can offer resources in exchange for others and they can also reply to offers made by other players. Our experimental results report that the DRL-based learnt policies significantly outperformed several baselines including random, rule-based, and supervised-based behaviours.  The DRL-based policy has a 53\% win rate versus 3 automated players (`bots'), whereas a supervised player trained on a dialogue corpus  in this setting achieved only 27\%,  versus the same 3 bots. This result supports the claim that DRL is a promising framework for training dialogue systems, and strategic agents with negotiation abilities.
\end{abstract}

\section{Introduction} %1 page
Artificially intelligent agents can require strategic conversational skills to negotiate during their interactions with other natural or artificial agents, e.g.\ {\it ``A: I will give/tell you X if you give/tell me Y?, B: Okay''}. While typical conversations of artificial agents assume cooperative behaviour from partner conversants, strategic conversation does not assume full cooperation during the interaction between agents \cite{asher:lascarides:2013}. Throughout this paper, we will use a strategic card-trading board game to illustrate our approach. Board games with trading aspects aim not only at entertaining people, but also at training them with trading skills. Popular board games of this kind include %Imperial 2030, Chicago Express, Acquire, Panic on Wall Street!, Airlines Europe, Crude, 
Last Will, Settlers of Catan, and Power Grid, among others \cite{TopBoardGames}. While these games can be played between humans, they can also be played between computers and humans. The trading behaviours of AI agents in computer games are usually based on  carefully tuned rules \cite{ThomasH02}, search algorithms such as Monte-Carlo tree search \cite{Szita2009,DobreLascarides2015}, and reinforcement learning with tabular representations \cite{georgila,Efstathiou:2014wd} or linear function approximation \cite{Pfeiffer2014,PapangelisGeorgila:2015sigdial}. However, the application of reinforcement learning is not trivial due to the complexity of the problem, e.g.\ large state-action spaces exhibited in strategic conversations. On the one hand, unique situations in the interaction can be described by a large number of variables (e.g.\ game board and resources available) so that enumerating them would result in very large state spaces. On the other hand, the action space can also be large due to the wide range of unique negotiations (e.g.\ givable and receivable resources). While one can aim for optimising the interaction via compression of the search space, it is usually not clear what features to incorporate in the state representation. This is a strong motivation for applying deep reinforcement learning for dialogue management, as first proposed by (anon citation), %OL: \cite{LemonEshghi2015}
so that the agent can simultaneously learn its feature representation and policy. In this paper, we present an application of deep reinforcement learning to learning trading dialogue for the game of Settlers of Catan.

%In addition, while one can optimise the whole game, one can also aim for a specialised solution. The latter is the focus of this paper by focusing on learning to trade only, rather than learning to play the whole game. 
%In addition, while previous work has focused on optimising negotiation strategies \cite{Pfeiffer2014,Szita2009}, our proposed approach focuses on learning human-like trading from human examples---despite the fact that in reality the ``best''  choice may not be the most human-like one, especially with non-expert player data. %automatic labelled data---bootstrapped from a small amount of labelled data.  

Our scenario for strategic conversation is the game of Settlers of Catan, where players take the role of settlers on the fictitious island of Catan---see Figure~\ref{integratedSystem}(left). The board game consists of 19 hexes randomly connected: 3 hills, 3 mountains, 4 forests, 4 pastures, 4 fields and 1 desert. In this island, hills produce {\it clay}, mountains produce {\it ore}, pastures produce {\it sheep}, fields produce {\it wheat}, forests produce {\it wood}, and the desert produces nothing. In our setting, four players attempt to settle on the island by building settlements and cities connected by roads. To build, players need specific resource cards, for example: a {\it road} requires clay and wood; a {\it settlement} requires clay, sheep, wheat and wood; a {\it city} requires three clay cards and two wheat cards; and a {\it development card} requires clay, sheep and wheat. Each player gets points for example by building a settlement (1 point) or a city (2 points), or by obtaining victory point cards (1 point each). A game consists of a sequence of turns, and each game turn starts with the roll of a die that can make the players obtain resources (depending on the number rolled and resources on the board). The player in turn can trade resources with the bank or through dialogue with other players, and can make use of available resources to build roads, settlements or cities. This game is highly strategic because players often face decisions about when to trade, what resources to request, and what resources to give away---which are influenced by what they need to build. A player can extend build-ups on locations connected to existing pieces, i.e.\ road, settlement or city, and all settlements and cities must be separated by at least 2 roads. The first player to win 10 victory points wins and all others lose.\footnote{\url{www.catan.com/service/game-rules}}

%\begin{figure}[t]
%  \begin{center}
%    \includegraphics[width=0.49\textwidth]{pics/Board-JSettlers.png}
%\caption{\label{JSettlers} Example board of the game ``Settlers of Catan'' \cite{ThomasH02}. The top-middle dialogue box is a chat interface that displays the game history---including trading offers and responses from all players}
%  \end{center}
%\end{figure}

In this paper, we extend previous work on strategic conversation that has applied supervised or reinforcement learning in that we simultaneously learn the feature representation and dialogue policy by using Deep Reinforcement Learning (DRL). We compare our learnt policies against random, rule-based and supervised baselines, and show that  the DRL-based agents perform significantly better than the baselines.

%\section{Case Study: The Settlers of Catan Game} % 1 page
%\section{Background on Deep Reinforcement Learning} %1 page
\section{Background}
\label{DRL}
%We model strategic conversation as a sequential decision-making problem using the Deep Reinforcement Learning (DRL) framework. 
A Reinforcement Learning (RL) agent learns its behaviour
from interaction with an environment and the physical or virtual agents within it, where situations are mapped to actions by maximizing a long-term reward signal \cite{Sutton_Barto-1998,Szepesvari:2010}. An RL agent is typically  characterized by: (i) a finite or infinite set of states $S=\{s_i\}$; (ii) a finite or infinite set of actions $A=\{a_j\}$; (iii) a stochastic state transition function $T(s,a,s')$ that specifies the next state $s'$ given the current state $s$ and action $a$; (iv) a reward function $R(s,a,s')$ that specifies the reward given to the agent for choosing action $a$ when the environment makes a transition from state $s$ to state $s'$; and (v) a policy $\pi:S \rightarrow A$ that defines a mapping from states to actions. %; and finally, (5) a neural network with weights $\theta$ that represents the state and action space jointly. 
The goal of an RL agent is to select actions by maximising its cumulative discounted reward defined as $Q^*(s,a)=\max_\pi \mathbb{E}[r_t+\gamma r_{t+1}+\gamma^2 r_{t+1}+...|s_t=s,a_t=a,\pi]$, 
%\begin{equation}
%Q^*(s,a)=\max_\pi \mathbb{E}[r_t+\gamma r_{t+1}+\gamma^2 r_{t+1}+...|s_t=s,a_t=a,\pi],
%\end{equation}
where function $Q^*$ represents the maximum sum of rewards $r_t$ discounted by factor $\gamma$ at each time step. While the RL  agent takes actions with probability $Pr(a|s)$ during training, it takes the best actions $\max_a Pr(a|s)$ at test time.
%SK where do the probabilities Pr(a|s) come from? Derived from the Q-function?

To induce the $Q$ function above we use Deep Reinforcement Learning as in \cite{mnih-dqn-2015}, which approximates $Q^*$ using a multilayer convolutional neural network. The $Q$ function of a DRL agent is parameterised as $Q(s,a;\theta_i)$, where $\theta_i$ are the parameters (weights) of the neural net at iteration $i$. More specifically, training a DRL agent requires a dataset of experiences $D=\{e_1,...e_N\}$ (also referred to as `experience replay memory'), where every experience is described as a tuple  $e_t=(s_t,a_t,r_t,s_{t+1})$. Inducing the $Q$ function consists in applying Q-learning updates over minibatches of experience $MB=\{(s,a,r,s')\sim U(D)\}$ drawn uniformly at random from the full dataset $D$. 
%SK D should be D_t ?
% OL no i don't think so. The D_t are the experience replays drawn from the full dataset D.
%SK okay, in that case, D should be defined as the full dataset (edited)

A Q-learning update at iteration $i$ is thus defined as the loss function $L_i(\theta_i)=\mathbb{E}_{MB} \left[ (r+\gamma \max_{a'} Q(s',a';\overline{\theta}_i)-Q(s,a;\theta_i))^2 \right]$,
%\begin{equation}
%%{(s,a,r,s')\sim U(D)}
%L_i(\theta_i)=\mathbb{E}_{MB} \left[ (r+\gamma \max_{a'} Q(s',a';\overline{\theta}_i)-Q(s,a;\theta_i))^2 \right],
%\end{equation}  
where $\theta_i$ are the parameters of the neural net at iteration $i$, and $\overline{\theta}_i$ are the target parameters of the neural net at iteration $i$. The latter are only updated every $C$ steps. This process is implemented in the learning algorithm {\it Deep Q-Learning with Experience Replay} described in \cite{mnih-dqn-2015}. 

%\newpage
\section{Policy Learning for Strategic Interaction} % 1 page
Our approach for strategic interaction optimises two tasks jointly: learning to offer and learning to reply to offers. In addition, our approach learns from constrained search spaces rather than unconstrained ones, resulting in quicker learning and also in learning from only legal (allowed) decisions.

\subsection{Learning to offer and to reply}
%SK I would prefer the terms 'trade' and 'trade offer' instead of 'trade negotiation' in the paragraph below; a trade negotiation refers to a negotiation dialogue 
%OL ok - agree - edited
A strategic agent has to offer a trade to its opponent agents (or players). In the case of the game of Settlers of Catan, an example trading  offer is {\it I will give anyone sheep for clay}. Several things can be observed from this simple example. First, note that this offer may include multiple givable and receivable resources. Second, note that the offer is addressed to all opponents (as opposed to one opponent in particular, which could also be possible). Third, note that not all offers are allowed at a particular point in the game --  they depend on the particular state of the game and resources available to the player for trading. The goal of the agent is to learn to make  legal offers that will yield the largest pay-off in the long run. 

A strategic agent also has to reply to  trading offers made by an opponent. In the case of the game of Settlers of Catan, the responses can be narrowed down to (a) accepting the offer, (b) rejecting it, or (c) replying with a counteroffer (e.g.\ {\it I want two sheep for one clay}). Note that this set of responses is available at any point in the game once there is an offer made by any agent (or player). Similarly to the task above, the goal of the agent is to learn to choose a response that will yield the largest pay-off in the long run.

While one can aim for optimising only one of the tasks above, a joint optimisation of the these two tasks equips an automatic trading agent with more completeness. To do that, given an environment state space $S=\{s_i\}$, trading negotiations $A^o$, and responses $A^r$, the goal of a strategic learning agent consists of inducing an optimal policy so that  action selection can be defined as $\pi^*(s) = \arg \max Q^*_{a \in \{A^o \cup A^r\}}(s,a)$,
%\begin{equation}
%a^* = \arg \max Q^*_{a \in \{A^o \cup A^r\}}(s,a),
%\end{equation}
where the $Q$ function is estimated as described in the previous section, $A^o$ is the set of trading negotiations in turn, and $A^r$ is the  set of responses.

\subsection{Deep Learning from constrained action sets}
While the behaviour of a strategic agent can be trained as described above, using deep learning with large action sets can be prohibitively expensive in terms of computation time. Our solution to this limitation consists in learning from constrained action sets rather than whole and static action sets. 
We distinguish two action sets, an action set $A^r$ which contains responses to trading negotiations and remains static, and an action set  $A^o$ which contains  those trading negotiations that are valid at any given point in the game (i.e.\ which the player is able to make due to the resources that they hold).  We refer to the latter action set as $\bar{A}^o$, which contains a dynamic set $|\bar{A}^o|\le|A^o|$ 
%SK $|\bar{A}^o|\subset|A^o|$ is more precise
of trading negotiations available according to the game state and available resources (e.g.\ the agent would not offer a particular resource if it does not have it). Thus, we reformulate the goal of a strategic learning agent as inducing an optimal policy so that action selection can be defined as $\pi^*(s) = \arg \max Q^*_{a \in {\bar{A}^o \cup A^r}}(s,a)$, 
%\begin{equation}
%a^* = \arg \max Q^*_{a \in {\bar{A}^o \cup A^r}}(s,a),
%\end{equation}
where the $Q$ function is still estimated as described in Section~\ref{DRL}, $\bar{A}^o$ is the constrained set of trading negotiations in turn (i.e.\ legal offers), and $A^r$ is the set of responses. Note that the size of $\bar{A}^o$ will vary depending on the game state.

%\newpage
\section{Experiments and Results} %1page
In this section we apply the approach above to conversational agents that learn to offer and to reply in the game of Settlers of Catan.

\subsection{Experimental Setting}

\begin{figure*}[th]
  \begin{center}
    \includegraphics[width=0.95\textwidth]{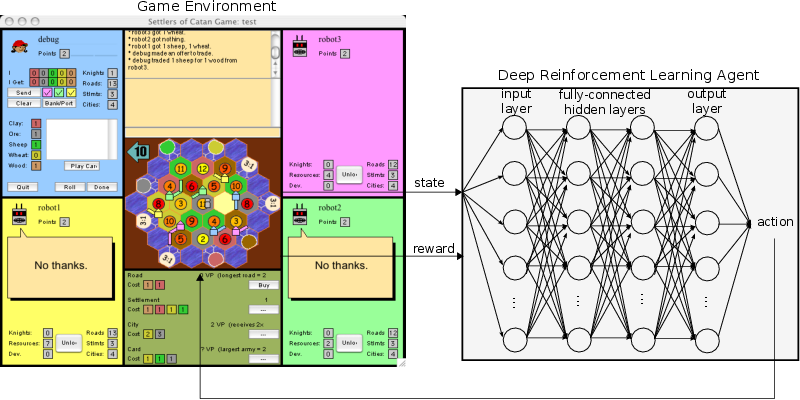}
\caption{\label{integratedSystem} Integrated system of the Deep Reinforcement Learning (DRL) agent for strategic interaction. (left) GUI of the board game ``Settlers of Catan'' \cite{ThomasH02}. (right) Multilayer neural network of the DRL agent--see text for details.}
  \end{center}
\end{figure*}

\subsubsection{Integrated learning environment}
%JSettlers, ConvNetJS, picture of integration
Figure~\ref{integratedSystem}(left) shows our integrated learning environment. On the left-hand side, the JSettlers benchmark framework \cite{ThomasH02} receives an action (trading offer or response) and outputs the next game state and numerical reward. On the right-hand side, a Deep Reinforcement Learning (DRL) agent receives the state and reward, updates its policy during learning, and outputs an action following its learnt policy. Our integrated system is based on a multi-threaded implementation, where each player makes use of a synchronised thread. In addition, this system runs under a client-server architecture, where  the learning agent acts as the `server' and the game acts as the `client'. They communicate by exchanging messages, where the server tells the client the action to execute, and the client tells the server the game state and reward observed. Our DRL agents are based on the ConvNetJS tool \cite{ConvNetJS}, which implements the algorithm `Deep Q-Learning with experience replay' proposed by \cite{mnih-dqn-2015}. We extended this tool to support multi-threaded and client-server processing with constrained search spaces.\footnote{The code of this substantial extension with an illustrative dialogue system is available at the following link: \url{https://github.com/cuayahuitl/SimpleDS}
} 

\begin{table}[t]
\footnotesize
\centering
\begin{tabular}{|c|l|l|l|}
\hline
\bf Num. & \bf Feature & \bf Domain & \bf Description \\ 
\hline 
%resources+"|"+hexes+"|"+nodes+"|"+edges+"|"+robber+"|"+turns
1 & hasClay & \{0...10\} & Number of clay units available\\
1 & hasOre & \{0...10\} & Number of ore units available\\
1 & hasSheep & \{0...10\} & Number of sheep units available\\
1 & hasWheat & \{0...10\} & Number of wheat units available\\
1 & hasWood & \{0...10\} & Number of wood units available\\
19 & hexes & \{0...5\} & Type of resource: 0=desert,1=clay,2=ore, 3=sheep,4=wheat,5=wood\\
54 & nodes & \{0...4\} & Where builds are located: 0=no settlement or city,\\
   &       &           & 1 and 2=opponent builds, 3 and 4=agent builds\\
80 & edges & \{0...2\} & Where roads are located:\\
   &       &           & 0=no road in given edge, 1=opponent road, 2=agent road\\
1 & robber & \{0...5\} & On type of resource: 0=desert,1=clay,2=ore, 3=sheep,4=wheat,5=wood\\
1 & turns & \{0..100\} & Number of turns of the game so far\\ 
\hline
\end{tabular}
\caption{Feature set (size=160) of the DRL agent for trading in the game of Settlers of Catan}\label{features}
\end{table}
%SK it would be useful to also include the total size of the state space
%SK 11^5 x (#allocations of 4 sheep/wheat/wood and 3 clay/ore hexes to 19 positions) x 6^19 x 4^54 x 3^80 x 6 x 101
%SK a huge number, although many combinations will never occur in practice

\subsubsection{Characterisation of the learning agent}
%states (table), actions, transition function, reward function (eq), learning algorithm and learning parameters
The {\it state space} $S=\{s_i\}$ of our learning agent includes 160 non-binary features that describe the game board and the available resources. Table~\ref{features} describes the state variables that represent the input nodes, which we normalise to the range [0..1]. These features represent a high-dimensional state space---only approachable via reinforcement learning with function approximation. 
%Given that these features are treated as non-discrete variables, they represent a high-dimensional continuous state space. 
%OL not sure it is really continuous? it is discrete values mapped in to [0...1]

The {\it action space} $A=\{a_i\}$ of our learning agents includes 70 actions for offering trading negotiations\footnote{Trading negotiation actions, where $C$=clay, $O$=ore, $S$=sheep, $W$=wheat, and $D=wood$: C4D, C4O, C4S, C4W, CC4D, CC4O, CC4S, CC4W, CD4O, CD4S, CD4W, CO4D, CO4S, CO4W, CS4D, CS4O, CS4W, CW4D, CW4O, CW4S, D4C, D4O, D4S, D4W, DD4C, DD4O, DD4S, DD4W, O4C, O4D, O4S, O4W, OD4C, OD4S, OD4W, OO4C, OO4D, OO4S, OO4W, OS4C, OS4D, OS4W, OW4C, OW4D, OW4S, S4C, S4D, S4O, S4W, SD4C, SD4O, SD4W, SS4C, SS4D, SS4O, SS4W, SW4C, SW4D, SW4O, W4C, W4D, W4O, W4S, WD4C, WD4O, WD4S, WW4C, WW4D, WW4O, WW4S. Example trade: C4D=clay for wood.} and 3 actions\footnote{Reply actions: accept, reject and counteroffer} for replying to offers from opponents. Notice that our offer actions only make use of up to two givable resources and only one receivable resource is considered. 
%This action space of 73 actions can be considered as fairly large.
%\footnote{Aote that a larger set of giveables and receivables quickly grows this size. Although this could be feasible within the DRL framework above, training policies with much 
% and therefore learning would become much slower. 

The {\it state transition function} of our agents is based on the game itself using the JSettlers framework \cite{ThomasH02}. In addition, our strategic interactions were carried out at the semantic level rather than at the word level, for example: {\it S4C} is a higher-level representation of {``I will give you sheep for clay''}. Furthermore, our trained agents were active only during the selection of trading offers and reply to offers, the functionality of the rest of the game was based
on the JSettlers framework.

The {\it reward function} of our agent is based on the game points provided by the JSettlers framework, but we make a distinction between reply actions and offer actions. This is due to the fact that we consider reply actions as high-level actions, and offer actions as lower-level ones. Our reward function is defined as:
\begin{equation}\nonumber
r=\left\{ \begin{array}{l} 
GainedPoints \times w_{gp} \mbox{ if }GainedPoints>0\\
TotalPoints \times w_{tp} \mbox{ otherwise},
\end{array} \right.
\end{equation}
where $GainedPoints$=points at time $t$ minus the points at time $t-1$, and $TotalPoints$ refers to the accumulated number of points of the trained agent during the game. We used the following weights for reply actions: $\{w_{gp}=1,w_{tp}=0.1\}$, and the following for offer actions: $\{w_{gp}=0.1,w_{tp}=0.01\}$. 
 
The {\it model architecture} consists of a fully-connected multilayer neural network with 160 nodes in the input layer (see Table~\ref{features}), 50 nodes in the first hidden layer, 50 nodes in the second hidden layer, and 73 nodes (action set) in the output layer. The hidden layers use RELU (Rectified Linear Units) activation functions to normalise their weights, see \cite{NairH10} for details. Finally, the learning parameters are as follows: experience replay size=30K, discount factor=0.7, minimum epsilon=0.05, learning rate=0.001, and batch size=64. A comprehensive analysis comparing multiple state representations, action sets, reward functions and learning parameters is left for future work.

\subsection{Experimental Results}
%learning curve, evaluation metrics, evaluation results, analysis, cross evaluation
We use the following baselines to compare our trained strategic agents, where we only switch the trading offers and reply behaviours---the remaining behaviour of the game remains constant and is provided by the JSettlers framework:
\begin{itemize}[leftmargin=*] 
\item {\bf Ran}: This agent chooses trading negotiation offers randomly, and replies to offers from opponents also in a random fashion. Although this is a weak baseline, we use it to analyse the impact of policies trained (and tested) against random behaviour.  %The remaining behaviour is given by the JSettlers framework.
\item {\bf Heu}: This agent chooses trading negotiation offers and replies to offers from opponents as dictated by the heuristic bots included in the JSettlers framework\footnote{The baseline trading agent referred to as `heuristic' included the following parameters, see \cite{GuheL14}: TRY\_N\_BEST\_BUILD\_PLANS:0, FAVOUR\_DEV\_CARDS:-5.}, see \cite{Thomas:2003vd,GuheL14} for details.
\item {\bf Sup}: This agent chooses trading negotiation offers using a random forest classifier \cite{Breiman01,HastieElAl2009}, and replies to offers from opponents using the heuristic behaviour above. This agent was trained from 32 games played between 56 different human players---labelled by multiple annotators. We compute the probability distribution of a human-like trade as $P(givable|evidence)=\frac{1}{Z}\prod_{b \in B} P_b(givable|evidence)$, where $givable$ refers to the class prediction (in our case, the givable resource), $evidence$ refers to observed features\footnote{Evidence: Number of resources available, number of builds (roads, settlements and cities), and the resource received.}, $P_b(.|.)$ is the posterior distribution of the $b$th tree, and $Z$ is a normalisation constant \cite{CriminisiSK12}. This classifier used 100 decision trees. Assuming that $Y$ is a set of givables at a particular point in time in the game, extracting the most human-like trading offer (givable $y^*$) given collected evidence (context of the game), is defined as $y^*=\arg\max_{y \in Y} Pr(y|evidence)$. The classification accuracy of this statistical classifier was 65.7\%---according to a 10-fold cross-validation evaluation \cite{CuayahuitlEtAl2015cgw,CuayahuitlEtAl2015ictai}. %(anonymised reference).
\end{itemize}

\begin{figure}[t]
  \begin{center}
    \includegraphics[width=0.80\textwidth]{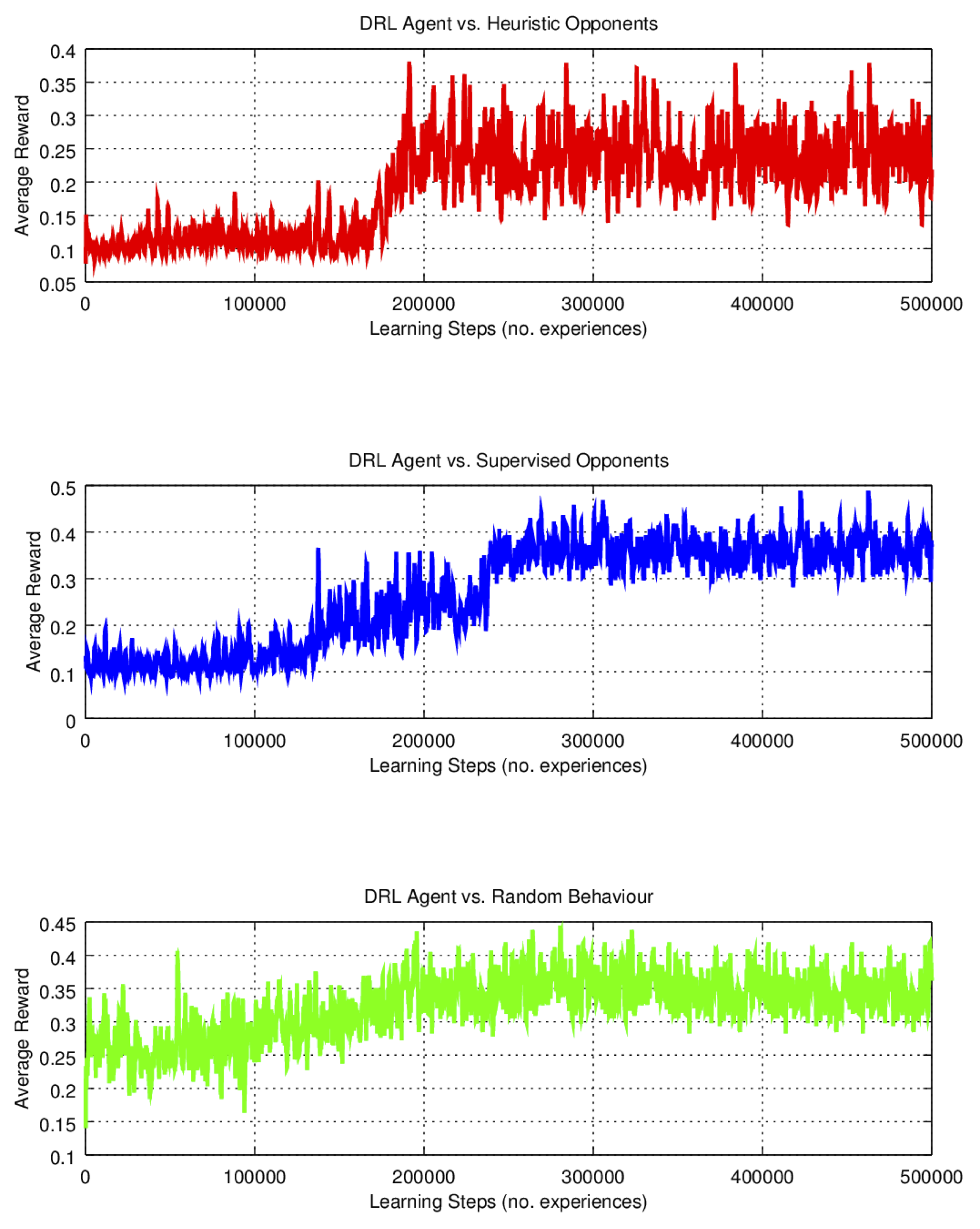}
\caption{\label{LearningCurves} Learning curves of Deep Reinforcement Learners (DRLs) against random, heuristic and supervised opponents. It can be observed that DRL agents can learn from different types of opponents---even from randomly behaving ones.}
  \end{center}
\end{figure}

\begin{table*}[t]
\footnotesize
\centering
\begin{tabular}{|l|c|c|c|c|c|c|c|c|c|}
\hline
\bf Comparison & \bf Winning & \bf Victory & \bf Offers & \bf Successful & \bf Total & \bf Pieces & \bf Cards & \bf Turns\\ 
\bf   between Agents & \bf Rate(\%) & \bf Points & \bf Made & \bf Offers & \bf Trades & \bf Built & \bf Bought & \bf p/Game\\
\hline 
\hline 
1 Ran vs. 3 Heu & 00.01 & 2.58 & 133.72  &  122.69 & 140.58 & 1.76 & 0.73 & 56.35\\
1 Ran vs. 3 Sup & 00.01 & 2.74 & 143.19 & 131.28 & 150.97 & 2.31 & 0.73 & 57.98 \\
1 Heu vs. 3 Ran & 98.46 & 10.15 & 41.63 & 18.19 & 167.12 & 13.81 & 0.24 & 45.59 \\
1 Heu vs. 3 Heu & {\bf 25.24} & 6.46 &   149.74 & 140.17 & 282.33 & 8.48 & 0.29 & 62.25 \\
1 Sup vs. 3 Ran & 97.30 & 10.13 & 45.61 & 19.89 & 175.38 & 13.84 & 0.24 & 47.97 \\
1 Sup vs. 3 Heu & {\bf 27.36} & 6.48 & 144.53  &  134.72 & 269.64 & 8.44 & 0.30 & 62.26 \\
\hline 
1 DRL$^{ran}$ vs. 3 Ran & 98.31 & 10.16 & 38.52 & 17.08 & 203.19 & 13.98 & 0.22 & 45.34 \\
1 DRL$^{ran}$ vs. 3 Heu & 49.49 & 8.06 & 144.82 & 137.13 & 353.98 & 11.04 & 0.27 & 62.72 \\
1 DRL$^{ran}$ vs. 3 Sup & 39.64 & 7.51 & 154.00 & 146.18 & 364.64 & 10.36 & 0.29 & 62.62 \\
\hline 
1 DRL$^{heu}$ vs. 3 Ran & 98.23 & 10.17 & 38.54 & 16.98 & 194.68 & 13.85 & 0.23 & 44.35 \\
1 DRL$^{heu}$ vs. 3 Heu & {\bf 53.36} & 8.22 & 146.84 & 139.12 & 343.29 & 11.37 & 0.27 & 61.46 \\
1 DRL$^{heu}$ vs. 3 Sup & 41.97 & 7.65 & 157.28 & 149.26 & 355.88 & 10.59 & 0.30 & 62.04 \\
\hline 
1 DRL$^{sup}$ vs. 3 Ran & 98.52 & 10.15 & 38.26 & 16.80 & 193.31 & 13.81 & 0.23 & 43.88 \\
1 DRL$^{sup}$ vs. 3 Heu & 50.29 & 8.14 & 150.62 & 142.66 & 348.14 & 11.31 & 0.28 & 62.59 \\
1 DRL$^{sup}$ vs. 3 Sup & {\bf 41.58} & 7.64 & 156.37 & 147.90 & 356.27 & 10.70 & 0.30 & 62.73 \\
%\hline 
%Conditional Random Field$^{man}$ vs Rule-based & 23.31 & 6.20 & 141.39  & 131.89 & 7.96 \\
%Bayesian Network$^{man}$ vs Rule-based & 24.20 & 6.20  &  141.59  &  131.72 & 7.98 \\
%Random Forest vs$^{man}$ Rule-based & {\bf 27.62} & {\bf 6.54}   &   145.61  &  135.84 & {\bf 8.50} \\
%Random Forest vs$^{sup-}$ Rule-based & 25.25 & 6.33   &   144.99  &  135.33 & 8.24 \\
%\hline 
%Conditional Random Field$^{auto}$ vs Rule-based & 23.10 & 6.19 & 152.30  & 142.61 & 7.92 \\
%Bayesian Network$^{auto}$ vs Rule-based & 21.30 & 6.03  &  155.36  &  145.48 & 7.63 \\
\hline
\end{tabular}
\caption{Evaluation results comparing Deep Reinforcement Learners (DRL) vs.\ 3 baseline traders (random, heuristic, supervised). Columns 2-7, show {\it average} results---of the player at the most left---over 10K test games. Notation: DRL$^{ran}$=DRL agent trained vs. random behaviour, DRL$^{heu}$=DRL agent trained vs.\ heuristic opponents, and DRL$^{sup}$=DRL agent trained vs.\ supervised opponents.}\label{CrossEvaluation}
\end{table*}

We trained three DRL agents against random, heuristic and supervised opponents---see Figure~\ref{LearningCurves}, which used 500K training experiences (around 2000 games each learning curve).
We evaluate the learnt policies according to a cross-evaluation using the following metrics in terms of averages per game (using 10 thousand test games per comparison): win-rate, victory points, (successful) offers, total trades, pieces built, cards bought, and number of turns. Our observations of the cross-evaluation, reported in Table~\ref{CrossEvaluation}, are as follows: %First, the DRL agents acquire very competitive strategic behaviour in comparison to the other types of agents---they simply win substantially more than their opponents. The DRL agents outperform the baselines in the following  metrics: win-rate, victory points and total trades. Second, training a DRL agent in the environment where it will be tested is better than training and testing across environments. For example, DRL$^{heu}$ versus heuristic behaviour is better than DRL$^{sup}$ versus heuristic behaviour. Third, the DRL agents trade more than their opponents, i.e. they accept more offered trading negotiations. This suggest that knowing when to accept, reject or counter offer a trading negotiation is crucial for winning. Fourth, the DRL agents find the supervised agent harder to beat. This is because the supervised agent is the strongest baseline---best winning rate from the baseline agents. Fifth, the DRL agents are able to learn competitive behaviour, even training from random behaviour leads to policies almost as good as those trained with stronger opponents. This suggest that DRL agents for strategic interaction can be also be trained without highly skilled opponents. Note that the fact that the agent with random behaviour hardly wins any  games, suggest that sequential decision-making in this strategic board game is far from trivial.  

\begin{enumerate}[leftmargin=*]
\item The DRL agents acquire very competitive strategic behaviour in comparison to the other types of agents---they simply win substantially more than their opponents. While random behaviour is  easy to beat with over 98\% win-rate, the DRL agents achieve over 50\% of win-rate against heuristic opponents and over 40\% against supervised opponents. These results substantially outperform the heuristic and supervised agents which achieve less than 30\% of win-rate (at $p<0.05$ according to a two-tailed Wilcoxon-Signed Rank Test). 
%OL should we also report results using standard RL?
%SK Wilcoxon signed rank is a paired test, which I believe is inappropriate here

\item The DRL agents outperform the baselines not just in win-rates but also in other metrics such as average victory points, pieces built and total trades. The latter is more prominent, for example, while the heuristic and supervised agents achieve between 270 to 280 trades per game, the DRL agents compared against heuristic and supervised agents achieve between 340 and 360 trades. This means that the DRL agents tend to trade more than their opponents, i.e.\ they accept more offered trading negotiations.
%OL: why that they *accept* more? could also be that they *offer* more?
 These differences suggest that knowing when to accept, reject or counter offer a trading negotiation is crucial for winning.
\item Training a DRL agent in the environment where it will be tested is better than training and testing across environments. For example, DRL$^{heu}$ versus heuristic behaviour is better (53.4\%  win-rate) than DRL$^{sup}$ versus heuristic behaviour (50.3\% win-rate). However, our results report that DRL agents trained using randomly behaving opponents are almost as good as those trained with stronger opponents. This suggests that DRL agents for strategic interaction can be also be trained without highly skilled opponents, presumably by tracking their rewards over time. 
%\item The DRL agents trade more than their opponents, i.e. they accept more offered trading negotiations. This suggest that knowing when to accept, reject or counter offer a trading negotiation is crucial for winning. 
%This strategy was discovered by the learnt agents without being told explicitly, i.e. they learned this strategy based on their feedback of the environment.
\item The DRL agents find the supervised agent harder to beat. This is because the supervised agent is the strongest baseline, which achieves the best winning rate of the baseline agents. It can be noted that the DRL agents versus supervised behaviour make more offers and trade more than the DRL agents versus heuristic behaviour. We can infer from this result that 
%offering and trading more does not necessarily result in more winning. Instead, %OL removed
knowing when to offer and when to trade seem crucial for better winning rates.

% \item The DRL agents and our two stronger baselines beat random behaviour with far fewer offers, trades and turns than non-random behaviour. Thus, offering and trading too much and too little are not good strategies---a balance is required for achieving high winning rates. 
%OL - removed this point. WHat are "our 2 stronger baselines"? It contradicts the points made above.

\item  The fact that the agent with random behaviour hardly wins any games, suggests that sequential decision-making in this strategic   game is far from trivial. 
%\item 
\end{enumerate}

In summary, strategic dialogue agents trained with deep reinforcement learning have the potential to acquire highly competitive behaviour, not just from training against strong opponents but even from opponents with random behaviour. This result may help to reduce the resources (heuristics or labelled data) required for training future strategic agents.

\section{Related Work} %0.5 page
Reinforcement learning applied to strategic interaction includes the following. \cite{Tesauro95} proposes reinforcement learning with multilayer neural networks for training an agent to play the game of Backgammon. He finds that agents trained with such an approach are able to match and even beat human performance. \cite{Pfeiffer2014} proposes hierarchical reinforcement learning for automatic decision making on object-placing and trading actions in the game of Settlers of Catan. He incorporates built-in knowledge for learning the behaviours of the game quicker, and finds that the combination of learned and built-in knowledge is able to beat human players. 
\cite{Efstathiou:2014wd} used reinforcement learning in non-cooperative dialogue, and focus on a small 2-player trading problem with 3 resource types, but without using any real human dialogue data. This work showed that explicit manipulation moves (e.g.\ ``I really need sheep") can be used to win when playing against adversaries who are gullible (i.e.\ they believe such statements) but also against adversaries who can detect manipulation and can punish the player for being manipulative \cite{el-semdial2014}.
%More recently, \cite{Keizer:2015semdial} compare trained  policies against hand-crafted traders and supervised traders created from human players. They found that rather than training trading policies on hand-crafted rule-based heuristics, a more successful approach is to train trading policies from a supervised classifier trained from human examples.  
More recently, \cite{Keizer:2015semdial} designed an MDP model for selecting trade offers, trained and evaluated within the full jSettlers environment (4 players, 5 resource types).  In comparison to the DRL model, it had a much more restricted state-action space, leading to significant, but more modest improvements over supervised learning and hand-coded baselines.

Other related work has been carried out in the context of automated non-cooperative dialogue systems, where an agent may  act to satisfy its own goals rather than those of other participants \cite{georgila}.  The game-theoretic underpinnings of non-cooperative behaviour have  also been investigated \cite{asher:lascarides:2008:short}. 
% For example, it may be useful for  a dialogue agent not to be fully cooperative when trying to gather information from a human, or 
Such automated agents are of interest when trying to persuade, argue, or debate,  or in the area of believable characters in video games  and educational  simulations \cite{georgila,arkin13}. Another arena in which strategic conversational behaviour has been investigated is negotiation \cite{traum:sigdial08}, where hiding information (and even outright lying) can be advantageous. 

Recent work on deep learning applied to games include the following. \cite{MaddisonHSS14} train a deep convolutional network for the game of Go, but it is trained in a supervised fashion rather than trained to maximise a long-term reward as in this work. A closely related work to ours is a DRL agent for text-based games \cite{narasimhan-kulkarni-barzilay:2015:EMNLP}. Their states are based on words, their policies are induced using game-based rewards, and their actions are based on directions such as `go east/west/south/north'. Another closely related work to ours is DRL agents trained to play ATARI games \cite{mnih-atari-2013}. Their states are based on pixels from down-sampled images, their policies make use of game-based rewards, and their actions are based on joystick movements. In contrast to these previous works which are based on navigation commands, our agents are use  trading dialogue moves (e.g.\ `I will give you ore and sheep for clay', or `I accept/decline your offer'), which are essential behaviours for strategic interaction.

This paper extends the recent work above on training strategic  agents using reinforcement learning, which have either used small state-action spaces or focused on navigation commands rather than negotiation dialogue. The learning agents described in this paper use a high dimensional state representation (160 non-binary features) and a fairly large action space (73 actions) for learning strategic non-cooperative dialogue behaviour. To our knowledge, our results report the highest winning rates reported to date in the game of Settlers of Catan, see \cite{GuheL14,Keizer:2015semdial,DobreLascarides2015}. The comprehensive evaluation reported in the previous section is evidence to argue that deep reinforcement learning is a promising framework for training strategic interactive agents.

%Given the machine learning efforts  applied to strategic interactive games, other forms of learning remain to be explored. They include not only direct but also inverse reinforcement learning to learn from trial and error, semi-supervised learning to learn from labelled and unlabelled data, unsupervised learning to learn from unlabelled data, multi-agent systems to learn behaviours considering the strategies of opponents, transfer learning so that agents do not have to be trained from scratch, and active learning to learn to ask what to do in uncertain situations while playing the game, among others---see \cite{Fuernkranz00machinelearning,CuayahuitlEtAl2013mlis,OlivierLopes2014} for an overview. Another direction to explore in strategic games includes a combination of planning and learning, which has shown more promising results than either in isolation \cite{MaddisonHSS14,DobreLascarides2015}. A further direction to explore includes end-to-end statistical training of language understanding \cite{CadilhacABL13,CuayahuitlDHL13}, game behaviour, and language generation \cite{lemon:londial08,NLE:9719879,DethlefsHCL13} using a unified learning framework.

\section{Concluding Remarks} % 0.5 page
The contribution of this paper is the first application of Deep Reinforcement Learning (DRL) to optimising the behaviour of strategic conversational agents. Our learning agents are able to: (i) discover what trading negotiations to offer, (ii) discover when to accept, reject, or counteroffer; (iii) discover strategic behaviours based on constrained action sets---i.e.\ action selection from legal actions rather than from all of them; and (iv) learn highly competitive behaviour against different types of opponents. All of this is supported by a comprehensive evaluation of three DRL agents trained against three baselines (random, heuristic and supervised), which are analysed from a cross-evaluation perspective. Our experimental results report that all DRL agents substantially outperform all the baseline agents. Our results are evidence to argue that DRL is a promising framework for training the behaviour of complex strategic interactive agents.
%This result may help to reduce but not discard the demanding requirements (such as heuristics or labelled data) for training strategic conversational agents. 

Future work can for example carry out similar evaluations as above in other strategic environments, and can also extend the abilities of the agents with other strategic features \cite{Lin:2010} and forms of learning \cite{CuayahuitlEtAl2013mlis,OlivierLopes2014}. In addition, a comparison of different model architectures, training parameters and reward functions can be explored in future work. Last but not least, given that our learning agents trade at the semantic level, they can be extended with language understanding/generation abilities to communicate verbally \cite{lemon-semdial2008,NLE:9719879}.

%OL:  Acknowedgements in final version: ERC STAC project Grant no. 269427, ESPRC BABBLE project EP/M01553X/1
\section*{Acknowledgments}
Funding from the European Research Council (ERC) project ``STAC: Strategic Conversation'' no.\ 269427 is gratefully acknowledged, see \url{http://www.irit.fr/STAC/}. Funding from the  ESPRC,    project EP/M01553X/1 ``BABBLE'' is gratefully acknowledged, see \url{https://sites.google.com/site/hwinteractionlab/babble}.

{\footnotesize
\bibliographystyle{abbrv}
\bibliography{hc-drl-nips2015}
}

\end{document}